\documentclass[9pt,twocolumn,twoside]{osajnl}
\journal{ao} % Choose journal (ao, aop, josaa, josab, ol, optica, pr)

\setboolean{shortarticle}{false}
\usepackage{stfloats}

\title{Panoramic annular SLAM with loop closure and global optimization}

\author[1]{Hao Chen}
\author[1]{Weijian Hu}
\author[2]{Kailun Yang}
\author[3]{Jian Bai}
\author[1,*]{Kaiwei Wang}

\affil[1]{National Engineering Research Center of Optical Instrumentation, Zhejiang University, Hangzhou 310058, China}
\affil[2]{Institute for Anthropomatics and Robotics, Karlsruhe Institute of Technology, 76131 Karlsruhe, Germany}
\affil[3]{State Key Laboratory of Modern Optical Instrumentation, Zhejiang University, Hangzhou 310027, China}

\affil[*]{Corresponding author: wangkaiwei@zju.edu.cn}

\begin{abstract}
In this paper, we propose panoramic annular simultaneous localization and mapping (PA-SLAM), a visual SLAM system based on panoramic annular lens.
A hybrid point selection strategy is put forward in the tracking front-end, which ensures repeatability of keypoints and enables loop closure detection based on the bag-of-words approach.
Every detected loop candidate is verified geometrically and the $Sim(3)$ relative pose constraint is estimated to perform pose graph optimization and global bundle adjustment in the back-end.
A comprehensive set of experiments on real-world datasets demonstrates that the hybrid point selection strategy allows reliable loop closure detection, and the accumulated error and scale drift have been significantly reduced via global optimization, enabling PA-SLAM to reach state-of-the-art accuracy while maintaining high robustness and efficiency.
\end{abstract}

\setboolean{displaycopyright}{true}

\begin{document}

\maketitle

%%%%%%%%%%%%%%%%%%%%%%%%%%  body  %%%%%%%%%%%%%%%%%%%%%%%%%%
\section{Introduction}
Pose estimation is a prerequisite for many applications, e.g., self-driving cars, autonomous robots and augmented/virtual reality.
Various sensors can be utilized in pose estimation, such as GPS, IMU, LIDAR and camera.
Among them, camera is especially favored by researchers due to its small size, low cost and abundant perceived information.
Pose estimation using only the continuous images captured by a single camera is called monocular visual odometry (VO).

\begin{figure}[htb]
\centering\includegraphics{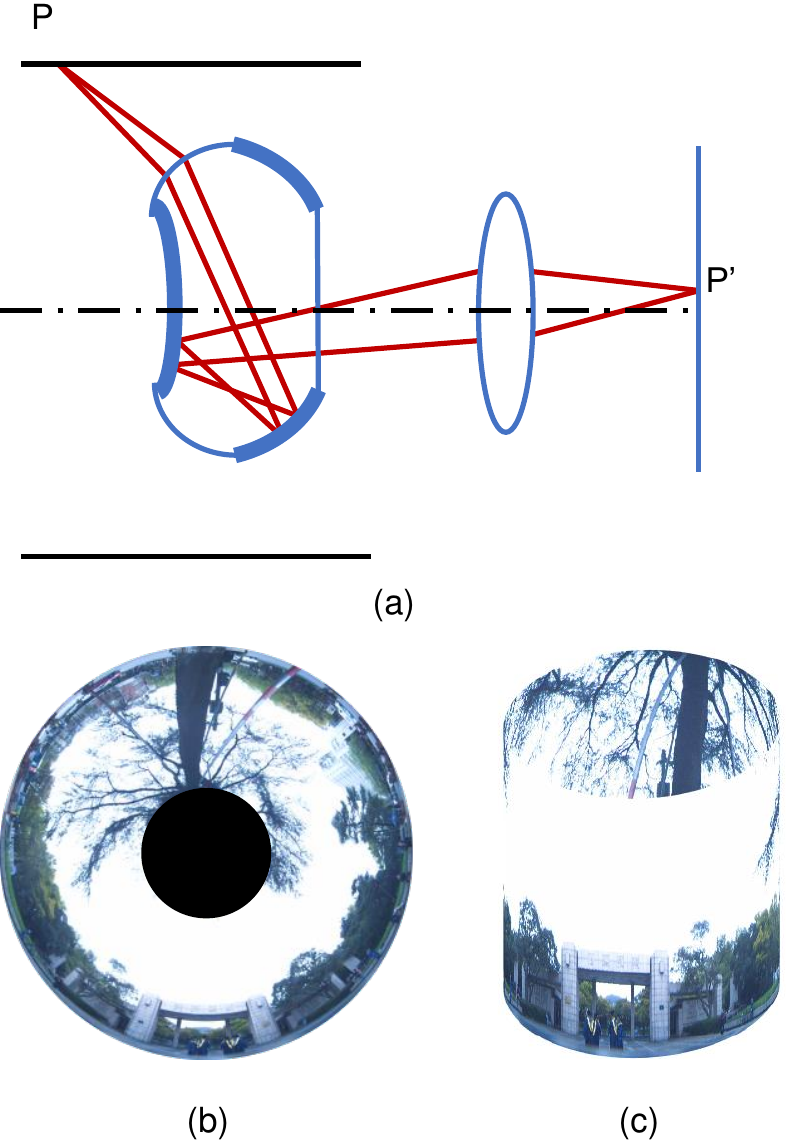}
\caption{(a) Schematic optical path in the PAL block. P and P' represent the object and image points, respectively. (b) A sample image captured by PAL camera. (c) The cylindrical object space of PAL.}
\label{fig:pal}
\end{figure}

A multitude of VO systems have been presented as of now, such as SVO \cite{forster2014svo} and DSO \cite{engel_direct_2017}. 
They are normally designed for the conventional pinhole cameras with a limited field of view (FOV).
The PALVO \cite{chen_palvo_2019} proposed in our previous work is a monocular VO based on panoramic annular lens (PAL). 
PAL can transform the cylindrical side view onto a planar annular image and obtain panoramic perception of $360^\circ$ FOV in a single shot \cite{luo2017compact}, as shown in Fig. \ref{fig:pal}.
Benefiting from the panoramic imaging, PALVO can handle some challenging scenarios that are difficult for conventional VO based on pinhole cameras.
For example, conventional VO will produce unreliable results when rotating with a fast angular velocity due to the rapid reduction of overlaps between adjacent frames,
and is greatly affected by dynamic components in the environment because of the limited FOV.
Compared with the traditional monocular VO, PALVO greatly improves the robustness of pose estimation in real application scenarios.

However, there still exits some problems in PALVO when it runs on a large scale and for a long time.
The first one is error accumulation \cite{fraundorfer2012visual}.
Since PALVO only maintains a local map consisting of the most recent several keyframes, the pose of each new frame is calculated by tracking the previous frame and the local map.
As a result, the errors introduced by each new frame-to-frame motion accumulate over time and cause the estimated trajectory to deviate from the actual path.
Secondly, it is impossible for PALVO to recover the absolute scale because only bearing information is available for a single camera, i.e., for a monocular VO, the motion and 3D map can only be recovered up to a scale factor.
But due to the inevitable errors of pose estimation, the scale of the motion estimated later may be distinct from that determined at the beginning,
which is known as scale drift \cite{strasdat2010scale}.
These problems will cause that although the camera actually revisits a certain place, it cannot be indicated from the estimated trajectory, i.e., the PALVO cannot ``close the loop''. 

To solve the above problems, we propose Panoramic Annular Simultaneous Localization And Mapping (PA-SLAM), which extends the previous PALVO by adapting it as the SLAM front-end to estimate camera poses with local localization consistency, and corrects error accumulation as well as scale drift with loop closure detection and global optimization in the back-end.

Compared with existing monocular visual SLAM systems based on pinhole camera with narrow FOV, the proposed PA-SLAM has the following advantages: 
Firstly, benefiting from the large FOV brought by PAL, PA-SLAM is less affected by dynamic components in the environment when performing loop closure detection.
For pinhole cameras, dynamic objects will have a significant influence on the image appearance, which will affect the loop closure detection \cite{yu2018ds}.
Secondly, the large FOV of PAL ensures that enough visual features can be extracted in a single shot, 
so pose estimation and loop closure detection will not be affected by the lack of features.
Thirdly, due to the cylindrical object space of PAL (Fig. \ref{fig:pal}(c)), loop closure detection in PA-SLAM is insensitive to travel direction, i.e., loop closure can be detected not only when the camera revisits a certain place in the same travel direction, but also in the perpendicular- and the reverse direction.
In contrast, pinhole cameras are mostly forward-looking and conventional visual SLAM can only detect loop closure in the same travel direction.

The contribution of this paper lies in threefold:
1. We present a method to extend sparse direct visual odometry (PALVO) to a full visual SLAM system;
2. A hybrid keypoint selection strategy is proposed to ensure repeatability of keypoints and to enable loop closure detection based on the bag-of-words approach, while maintaining high computational efficiency;
3. We verify the presented PA-SLAM on real-world datasets collected by a remote control vehicle equipped with a PAL camera. 
Several comparative experiments with existing VO/SLAM based on both panoramic and perspective images are conducted, demonstrating the superiority of the proposed PA-SLAM.

The remainder of the paper is organized as follows. 
Section 2 reviews the related work. 
Our algorithms are described in detail in Section 3.
In Section 4, extensive experiments are conducted to evaluate the proposed PA-SLAM.
Finally, we draw our conclusion in Section 5.

\section{Related work}

\subsection{Visual SLAM}

Many visual SLAM systems have been proposed during the last decade.
One of the most influential visual SLAM approaches is ORB-SLAM2 \cite{mur-artal_orb-slam2}.
It uses the same ORB (Oriented FAST and rotated BRIEF) \cite{rublee2011orb} features for tracking, mapping, and place recognition tasks.
A bag-of-words (BoW) \cite{galvez2012bags} place recognizer built on DBoW2 with ORB features is embedded for loop closure detection.
As a feature-based method, ORB-SLAM2 needs to extract ORB features on both keyframes and non-keyframes, and relies on feature matching to obtain data association, which is a time-consuming task.

Another famous visual SLAM is LSD-SLAM \cite{engel2014lsd}, which utilizes FAB-MAP \cite{glover2012openfabmap}, an appearance-based loop detection algorithm, to detect large-scale loop closures.
However, FAB-MAP needs to extract its own features, so none of information from the VO front-end can be reused in loop detection.
Besides, the relative pose calculation relies on direct image alignment, which means that all the images of past keyframes need to be kept in memory, resulting in large memory costs in long-time running.

Some researchers have also done some work to extend VO to SLAM. 
For example, LDSO \cite{gao_ldso} is extended by adding loop closure detection and pose map optimization to DSO. 
As a VO based on the direct method, DSO tracks the pixels with high gradient in the image through direct alignment in the front-end, and the back-end takes use of the sliding window method based on keyframes. 
LDSO proposed to gear point selection towards repeatable features,
which makes it possible to apply the BoW method similar to ORB-SLAM2 for loop closure detection, and estimate constraints using geometric techniques.
Similarly, VINS-Mono \cite{qin_vins-mono} also calculates additional feature point descriptors in keyframes and utilizes BoW for loop closure detection.
However, LDSO and VINS-Mono only conduct pose graph optimization, but do not perform the global bundle adjustment (BA).

Inspired by LDSO and VINS-Mono, we extract additional features and take use of BoW to detect loop closure. 
Compared to them, PA-SLAM has three main advantages:
(1) The extracted feature points are not all involved in tracking front-end, but only part of the feature points will be aggregated in the pose estimation and structure reconstruction, which enables reliable loop closure detection and meanwhile ensures the computational efficiency;
(2) The global BA can be carried out flexibly after pose graph optimization, further improving localization accuracy and global mapping consistency;
(3) The loop closure detection of PA-SLAM is direction-insensitive, while visual SLAM based on pinhole cameras can only handle the loop closure when traveling in the same direction.

\subsection{Panoramic visual localization}
\label{sec:pano_vpr}

In recent years, many researchers have been exploring the application of panoramic images in positioning tasks, including visual place recognition (VPR), VO and SLAM.

For the VPR task, Murillo and Josecka \cite{murillo2009experiments} proposed place recognition utilizing GIST descriptors, which has achieved satisfactory performance on large-scale datasets. 
Cheng et al. \cite{cheng2019panoramic} presented a panoramic image retrieval method based on NetVLAD \cite{arandjelovic2016netvlad} to tackle the challenges of various appearance variations between query and database images.
Oishi et al. \cite{oishi2019seqslam++} proposed to use panoramic images as one of the multi-modal data for robot localization and navigation, during which the panoramic images are matched using hand-crafted features and a sliding window scheme.

For VO and SLAM, some researchers have studied the advantages of large FOV. 
For example, SVO, DSO, VINS-Fusion, ORB-SLAM3 have been extended to support fisheye lenses \cite{forster2016svo, matsuki2018omnidirectional, qin2019b, campos2020orb}. 
Wang et al. \cite{cubemapslam} presented CubemapSLAM, which is a real-time feature-based SLAM system for fisheye cameras.
Lin et al. \cite{lin2018pvo} proposed PVO based on Ricoh Theta V panoramic camera, which is a multi-camera system composed of two fisheye lenses and produces 360$^\circ$ FOV through stitching images. 
Seok et al. presented ROVO \cite{seok2019rovo} and OmniSLAM \cite{won2020omnislam} for a wide-baseline multiview stereo setup with wide-FOV fisheye cameras.
Gutierrez et al. \cite{gutierrez2011adapting} developed a real-time EKF (extended Kalman filter) based visual SLAM system for catadioptric cameras. 
Compared to these works with wide-FOV imaging systems (fisheye lenses, catadioptric cameras
and multi-camera panoramic imaging systems), we exploit PAL in the proposed PA-SLAM, which has significant advantages of relative small distortions, single-shot panoramic perception and the compact structure \cite{huang2013stray}.
These advantages make PAL camera an ideal sensor for localization and perception tasks \cite{hu2019indoor, yang2019ds, fang2020cfvl}.

\begin{figure*}[htb]
\centering\includegraphics{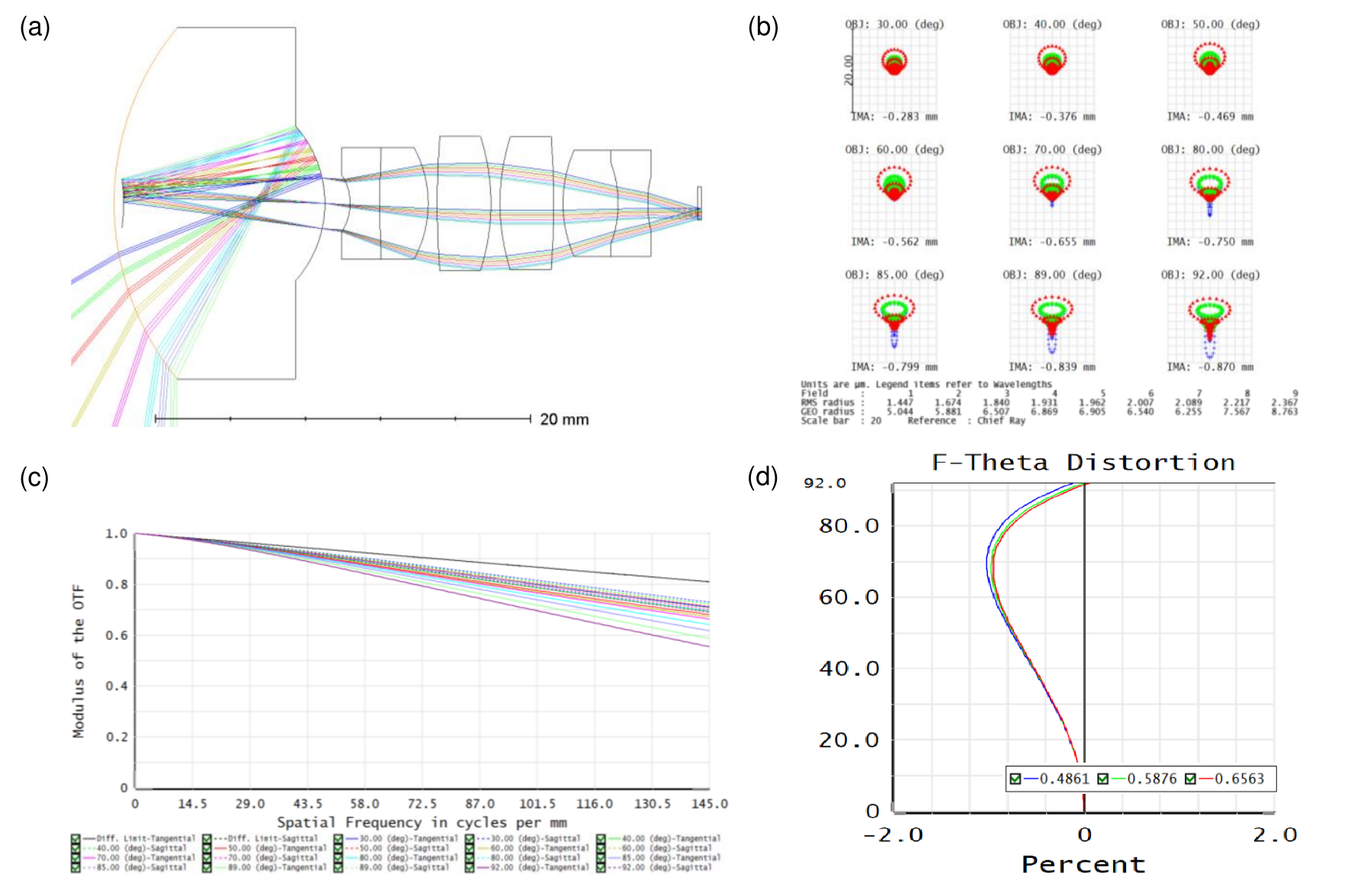}
\caption{(a) Structure of the PAL optical system. (b) Spot diagrams of $30^\circ$, $40^\circ$, $50^\circ$, $60^\circ$, $70^\circ$, $80^\circ$, $85^\circ$, $89^\circ$ and $92^\circ$ fields. (c) MTF of different fields below 145 lp/mm. The solid and dashed lines denote tangential and sagittal MTF, respectively. The black curve represents the MTF of the diffraction limit. (d) F-Theta distortion in different fields. }
\label{fig:optical_design}
\end{figure*}

\section{Optical design}

A self-designed PAL with a $360^\circ\times$($30^\circ$-$92^\circ$) FOV \cite{sun2019multimodal} is utilized in this paper, as well as a global shutter camera with a 2048$\times$2448 imaging resolution and 3.45 \textmu m pixel size.
The specifications of the optical system are listed in Table \ref{tab:optical_design}.
As indicated, the PAL system is designed for the working spectrum at 0.486-0.656 \textmu m with a F-number of 1.8.
The total length of the PAL system is 31.3 mm.

\begin{table}[h]
\centering
\caption{\bf Specifications of the PAL system.}
\label{tab:optical_design}
\begin{tabular}{lc}
\toprule
\textbf{Parameter}      & \textbf{Specification}                            \\
\midrule
Working spectrum        & 0.486-0.656 \textmu m                             \\
F$\setminus$\#          & 1.8                                               \\
FOV                     & $360^\circ\times$($30^\circ$-$92^\circ$)          \\
Total length            & 31.3 mm                                           \\
F-Theta distortion      & \textless{}1.0\%                                  \\
MTF                     & \textgreater{}0.55 at 145 pl/mm                   \\
Camera                  & 2048$\times$2448 with 3.45 \textmu m pixel size   \\ 
\bottomrule
\end{tabular}
\end{table}

The structure of the PAL optical system is shown in Fig. \ref{fig:optical_design}(a).
Generally, a PAL system consists of two components, the PAL block and a relay lens.
The PAL block is composed of two refractive and two reflective surfaces.
As the PAL block produces annular image mapping, a relay lens with a symmetrical structure can effectively balance aberrations and achieve adequate imaging quality.
Fig. \ref{fig:optical_design}(b) illustrates the spot diagram of the system.
As can be seen, the maximum RMS radius is 2.367 µm at the $92^\circ$ field, which is smaller than the pixel size of the camera and can therefore realise sharp imaging.
Fig. \ref{fig:optical_design}(c) is the MTF of the optical system.
With a camera pixel size of 3.45 \textmu m, the spatial cutoff frequency is 145 lp/mm.
As Fig. \ref{fig:optical_design}(c) indicates, the MTF below 145~lp/mm is above 0.55, meeting the resolution requirement of the camera we use.
Additionally, the F-Theta distortion sustains less than 1\% in all of the FOV (Fig. \ref{fig:optical_design}(d)), 
delivering competitive advantage compared to other wide-FOV imaging systems mentioned in Section \ref{sec:pano_vpr}.

\section{Algorithm}

Before going into PA-SLAM in more detail, we briefly review the pipeline of PALVO, which is the previous work of this paper. 

PALVO takes use of a sparse direct method, meaning that the feature correspondence is not explicitly calculated.
During the initialization process, feature points are tracked from frame to frame using Lucas-Kanade feature tracking (KLT) \cite{bouguet2001pyramidal}, and essential matrix is calculated to recover the poses and 3D map points of the first two keyframes. 
In the tracking thread, a coarse-to-fine strategy is adopted to estimate the camera pose for each new frame: 
Firstly, track the previous frame to obtain the coarse pose estimation through photometric error minimization; 
Secondly, track the local map by projecting keypoints to the current frame and optimizing the projection position;
Finally, the camera pose is fine-tuned by minimizing the reprojection error. 
In the mapping thread, a fixed-size local map is maintained, and the depth of keypoints in the local map are updated through a depth filter.
When the number of keyframes in the local map exceeds a threshold, the furthest keyframe will be discarded.

\begin{figure*}[htb]
\centering\includegraphics{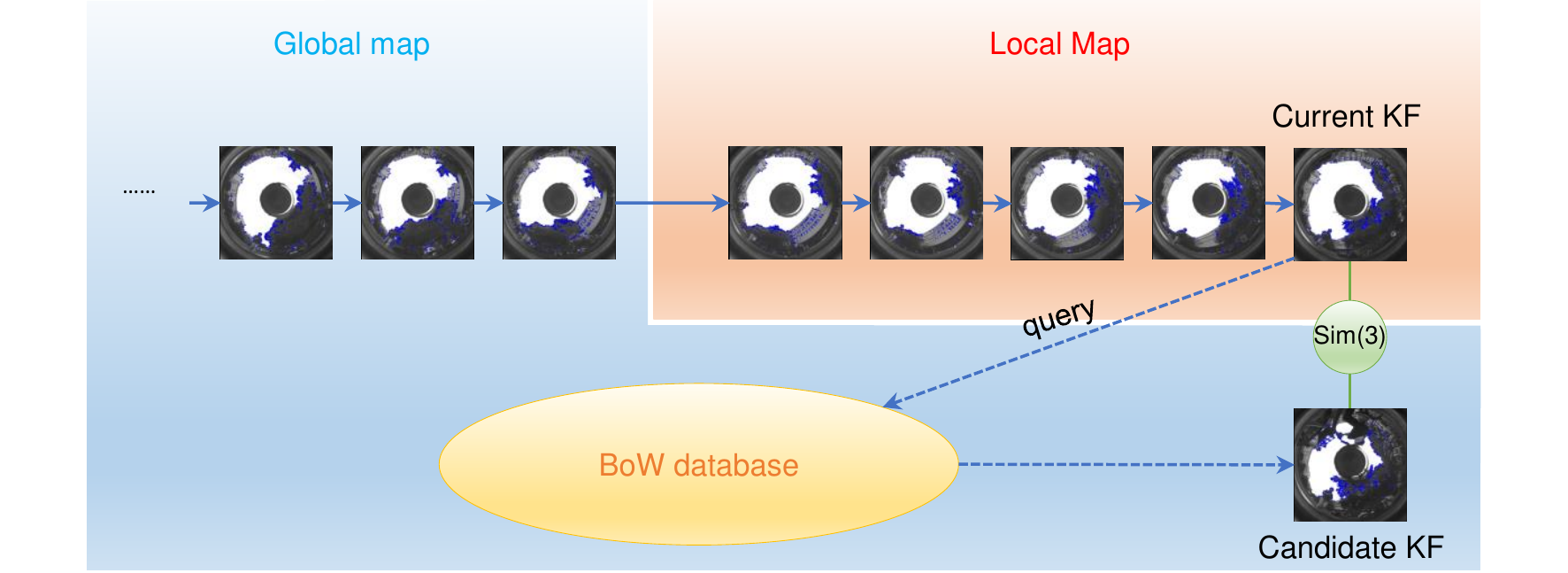}
\caption{Framework of PA-SLAM. Keyframes moved out from the local map are managed by the global map, and a BoW database is constructed. Loop candidates are proposed by querying the BoW database and verified geometrically. Once a loop closure is detected successfully, the $Sim(3)$ relative pose constraint between the candidate keyframe (KF) and the current keyframe will be calculated.}
\label{fig:algorithm}
\end{figure*}

In this paper, we adapt PALVO as the front-end of PA-SLAM to estimate frame-to-frame camera poses, and correct error accumulation as well as scale drift with loop closure detection and global optimization in the back-end.
The tracking and mapping threads are inherited from PALVO.
The difference lies in that each keyframe moved out from the local map is not simply discarded but added to the global map with a BoW database, as shown in Fig.~\ref{fig:algorithm}. 
The task of loop closure detection is carried out by querying the BoW database and the loop candidates are verified geometrically.
Once a loop closure is successfully detected, the $Sim(3)$ transformation (3D similarity transformation) between the candidate keyframe and the current keyframe is calculated and added to the pose graph as a constraint.
Then, all the poses of keyframes in the global map are adjusted by pose graph optimization and followed by global BA. 

\subsection{Selection of feature points}

As mentioned above, the front-end of PA-SLAM is a VO based on a sparse direct method, which features pose estimation via sparse image alignment rather than explicit feature matching.
There exists several open challenges in adapting such a direct visual odometry to reuse the existing map.
First of all, PALVO does not care about the repeatability of the tracked pixels (keypoints).
Thus, if we simply attempt to reuse the tracked keypoints in the front-end and compute descriptors for them, it is likely to result in poor loop closure detection.
Secondly, when the loop closure is detected and the inter-frame $Sim(3)$ transformation computation is carried out, the actual transformation matrix may be quite different from the unit matrix (the initial guess of optimization process). 
At this time, sparse image alignment will be invalid.

Therefore, we propose a hybrid point selection strategy in PA-SLAM.
When a frame is selected as a keyframe, new keypoints extraction will be carried out before it is sent into the depth filter.
The hybrid point selection strategy means that when extracting new keypoints, it is more inclined to consider ORB feature points, i.e., more ORB feature points are used as keypoints for tracking in the front-end.
In areas with insufficient features, pixels with a high gradient are used to supplement.
This strategy has the following advantages: 
Firstly, ORB feature points are actually FAST corners with good repeatability, and have been proved to be an effective feature for loop closure detection in visual SLAM; 
Secondly, once a loop closure is detected, feature matching can be easily obtained, which is convenient for geometric check and inter-frame $Sim(3)$ transformation computation.

\begin{figure*}[htb]
\centering\includegraphics{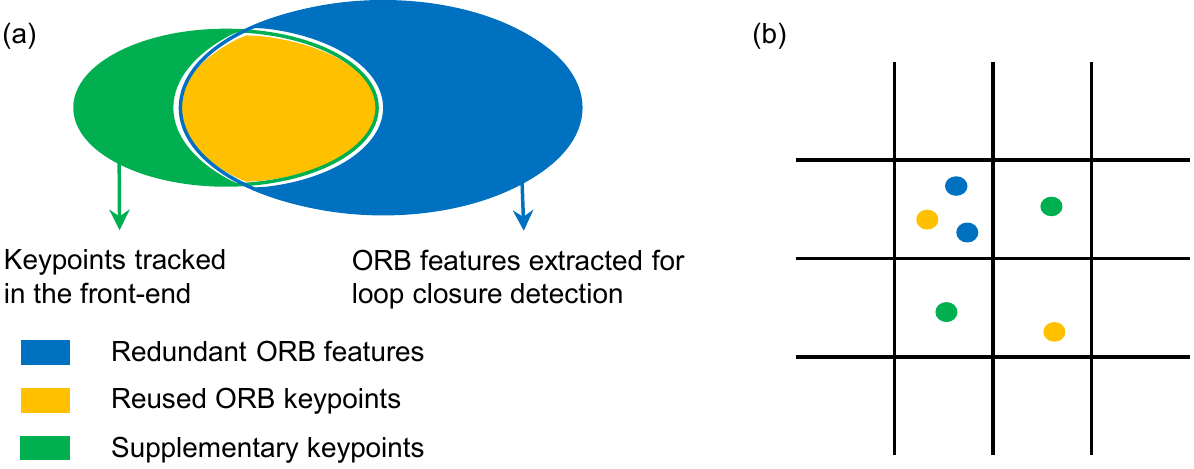}
\caption{The hybrid point selection strategy. 
(a) ORB features will be extracted from new keyframes for loop closure detection, but only a part of them are selected for tracking in the front-end. 
Some supplementary keypoints with high image gradient will also been involved in tracking if necessary. 
(b) Keypoint selection. 
The image is divided in a grid, and for each cell only the ORB keypoint with the highest Harris response is selected for tracking in the front-end. 
And for the cells without ORB feature points, the image gradient in the cell is computed and the pixel with the highest gradient is selected as a supplementary keypoint.}
\label{fig:point_selection}
\end{figure*}

\begin{figure*}[htb]
\centering\includegraphics{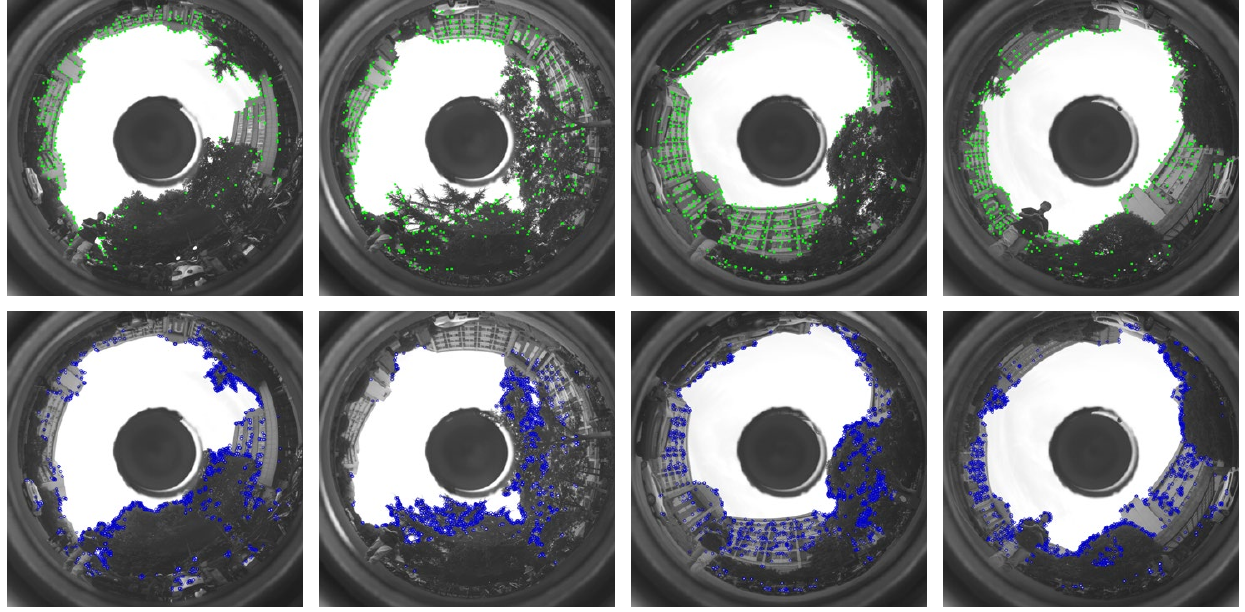}
\caption{Upper row: keypoints tracked in the front-end (drawn in green). Lower row: ORB features extracted for loop closure detection (drawn in blue).}
\label{fig:feat_component}
\end{figure*}

In the implementation, redundant ORB features will be extracted from new keyframes so as to ensure the performance of loop closure detection, and all the features are involved in generating BoW image descriptors, as shown in Fig. \ref{fig:point_selection}(a).
But not all ORB feature points are picked as depth filter seeds considering real-time performance.
The image is divided in a grid, and for each cell only the one with the highest Harris response is selected for depth recovery.
And for the cells without ORB feature points, the image gradient in the cell is computed and the pixel with the highest gradient is selected as a supplementary keypoint (same as the original strategy in PALVO) and fed into the depth filter, as shown in Fig. \ref{fig:point_selection}(b).
Fig. \ref{fig:feat_component} depicts the extracted ORB features for loop closure detection (lower row) and the tracked keypoints in the front-end (including reused ORB keypoints and the supplementary keypoints with a high image gradient, upper row) during one run.

\subsection{Loop closure detection and geometric check}

As mentioned above, redundant ORB features will be extracted from new keyframes and then DBoW3 \cite{dbow3} is utilized to transform ORB feature descriptors to BoW vectors and build a BoW database, and the database is queried to propose loop candidates for the current keyframe.
It is worth noting that the loop closure is only retrieved outside the local map, i.e., only the historical keyframes in the global map can be picked.

There may be false positives in loop closure detection via BoW database retrieval.
Therefore, a geometric check must be performed for each loop candidate. 
Here, geometric check is done via verifying epipolar constraints. 
For each pair of ideal matching feature points $\mathbf{u}_{r}$ and $\mathbf{u}_{c}$, it should be satisfied that

\begin{equation}
    \pi ^ {-1} \left( \mathbf{u}_{r} \right) ^ T
    \cdot \mathbf{E} \cdot
    \pi ^ {-1} \left( \mathbf{u}_{c} \right)
    = 0,
\end{equation}
where $\pi^{-1}\left(\cdot\right)$ is the back-projection function, $\mathbf{u}_{r}$ and $\mathbf{u}_{c}$ are the pixel coordinates of the matching ORB feature points on the reference frame (candidate) and the current keyframe respectively, and $\mathbf{E}$ is the essential matrix. 

Specifically, feature matching is first carried out between the candidate- and the current keyframe, and good matches are selected according to the matching distance.
Based on the good matches, the essential matrix is computed using the 8-point method \cite{longuet1981computer} with a random sample consensus (RANSAC) scheme \cite{derpanis2010overview}, and the number of inliers is counted. 
Only if the inlier number is greater than a threshold, geometric check is considered to be successful.

The same technology is also used in the initialization process of PALVO.
The difference lies in that KLT is used to obtain the correspondence between pixels in initialization, 
while ORB feature matching is used here. 
This is because the parallax between the loop candidate frame and the current keyframe may be large, so the optical flow can not be calculated effectively.

\subsection{$Sim(3)$ computation}

If a loop closure is successfully detected, the $Sim(3)$ relative pose from the loop candidate frame to the current keyframe will be calculated, where the 3D coordinates of the matching points are required. 
As mentioned above, not all extracted ORB feature points are fed into the depth filter to recover depth, so we can not guarantee that every matching point has its corresponding 3D coordinates.
In view of this, we propose an approximate strategy to obtain the depth of the feature points.

Specifically, for each feature point, if there exists a 3D map point in the same grid cell with it, the depth of this map point is regarded as the depth of the feature point.
If the opposite is true, then we search its 3$\times$3 neighborhood grid cells to find adjacent 3D map points and calculate the weighted average depth as the depth of the feature point.
After the depth of matching feature points are obtained, the 3D coordinates can be calculated using the back-projection function.

For the matching points with effective 3D coordinates, the algorithm proposed in \cite{horn_closed-form_1987} is utilized to solve $Sim(3)$. 
In order to ensure the robustness of the solution, RANSAC scheme is adopted.

\subsection{Pose graph optimization and global BA}

The $Sim(3)$ relative pose indicates the rotation, translation and scale constraints between the loop candidate frame and the current keyframe.
By adding this constraint during pose graph optimization, error accumulation and scale drift in this period of time can be reduced.

In general, the relative pose estimation between adjacent frames in the local map is reliable, 
but due to error accumulation and scale drift, the error of global pose gradually increases over time. 
Pose graph optimization is to optimize the pose of each keyframe with the constraints of the relative pose transformation between keyframes. 
Since the estimated pose in the front-end is $SE(3)$, it is upgraded to $Sim(3)$ during optimization so as to adjust its scale, and the initial scale is set to 1. The form of error in pose graph optimization is

\begin{equation}
\mathbf{e}_{i j} = { 
    \ln 
    \left( 
        \mathbf{S}_{i j} 
        \hat{\mathbf{S}}_{j}^{-1} 
        \hat{\mathbf{S}}_{i} 
    \right) 
}^\vee,
\end{equation}
where $S_i$ represents the $Sim(3)$ pose of the keyframe $i$, $S_ {ij}$ denotes the $Sim(3)$ relative pose between the keyframe $i$ and $j$, and $\hat{\ }$ denotes the estimated value of a variable.

After pose graph optimization, the global BA is then performed to fine-tune the 3D coordinates of all map points and poses of all keyframes in the global map by minimizing the reprojection error. The error term is

\begin{equation}
\mathbf{e}_{i}^{m} = 
    \mathbf{u}_{i}^{m} - 
    \pi \left( 
        \hat{\mathbf{T}}_{i} \cdot 
        \hat{\mathbf{P}}_{m} 
    \right),
\end{equation}
where $\mathbf{u}_{i}^{m}$ represents the observed projection of the 3D map point $m$ in the keyframe $i$,
$\pi\left(\cdot\right)$ is the projection function,
$\mathbf{T}_{i}$ is the pose of the keyframe $i$, and $\mathbf{P}_{m}$ is the 3D coordinate of the map point $m$.

It is also important to note that in order not to interfere with the pose estimation process in the front-end, the estimated poses of active keyframes in the local map are all fixed during pose graph optimization and global BA. 
Only the global poses of the old part of the trajectory will tend to be modified.
We utilize g2o, a graph optimization library proposed in \cite{grisetti2011g2o} for optimization tasks.

\section{Experiments}

\subsection{Experimental setup}

\begin{figure}[htb]
\centering\includegraphics{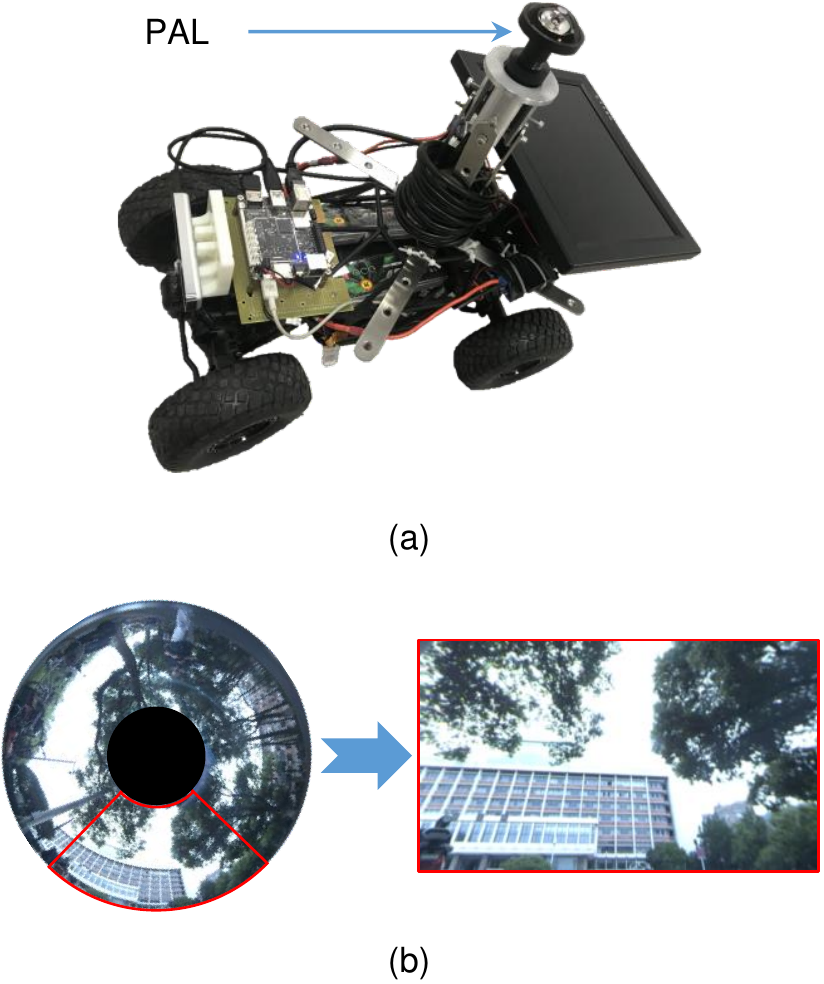}
\caption{(a) The remote control vehicle equipped with a PAL camera for dataset collection. (b) Perspective images used for comparative experiments are synthesized using a virtual pinhole camera and the reprojection method.}
\label{fig:experiment_setup}
\end{figure}

The PAL videos of real scenarios used in the following experiments are captured using a remote control vehicle equipped with the self-designed PAL camera, as shown in Fig. \ref{fig:experiment_setup}(a). The imaging resolution of the camera is 2048$\times$2448. For the sake of real-time performance, the image resolution is cropped and downsampled to 720$\times$720 before being fed into PA-SLAM system. 

In order to compare with SLAM systems based on the conventional pinhole camera, we use a virtual pinhole camera and the reprojection method to obtain perspective images.
This is perfectly feasible as the PAL imaging model follows a clear F-Theta law \cite{Zhou:16}.
As shown in Fig. \ref{fig:experiment_setup}(b), the PAL image is first back-projected into 3D space using a calibrated PAL camera model, and then re-projected into a perspective image using a virtual pinhole camera model with a $90^\circ$ horizontal FOV.  
In this way, PAL and perspective image sequences share the same FPS and timestamp, ensuring the fairness of the comparison to the maximum extent.

OmniCalib calibration tool \cite{scaramuzza2006toolbox} is used to calibrate the PAL camera. Evo \cite{grupp2017evo} and the method proposed in \cite{engel_photometrically_nodate} are used to evaluate the trajectory.

\subsection{Loop closure detection test}

\subsubsection{Relationship between feature number and loop closure detection}

In this section, the relationship between the performance of loop closure detection based on PAL images and the number of ORB features is studied. 
Videos captured by the remote control vehicle are used and the total length of the trajectory is about 500 meters. 
We select one image as a keyframe every fixed number of images (set to 30 in this paper), and take all the keyframes as the database to be queried.
Then for each query frame, we use the algorithm described in Section 3.2 for loop closure detection. 
For each detected loop closure candidate, if the difference of index between the candidate frame and the current query frame is less than the interval number of keyframes (30 in this paper), it is considered to be a true positive (TP) loop closure; 
Otherwise, it will be treated as false positive (FP). 
In addition, all the query frames that fail in detecting loop closure are defined as false negative (FN). 

The precision-recall curve is used to characterize the performance of the loop closure detection algorithm. 
Precision (P) and recall (R) can be calculated as follows:
\begin{equation}
    P = \frac{TP}{TP+FP},
\end{equation}
\begin{equation}
    R = \frac{TP}{TP+FN}.
\end{equation}
The higher the curve, the higher the recall at the same precision, which means the better the performance of the algorithm.

\begin{figure}[tb]
\centering\includegraphics{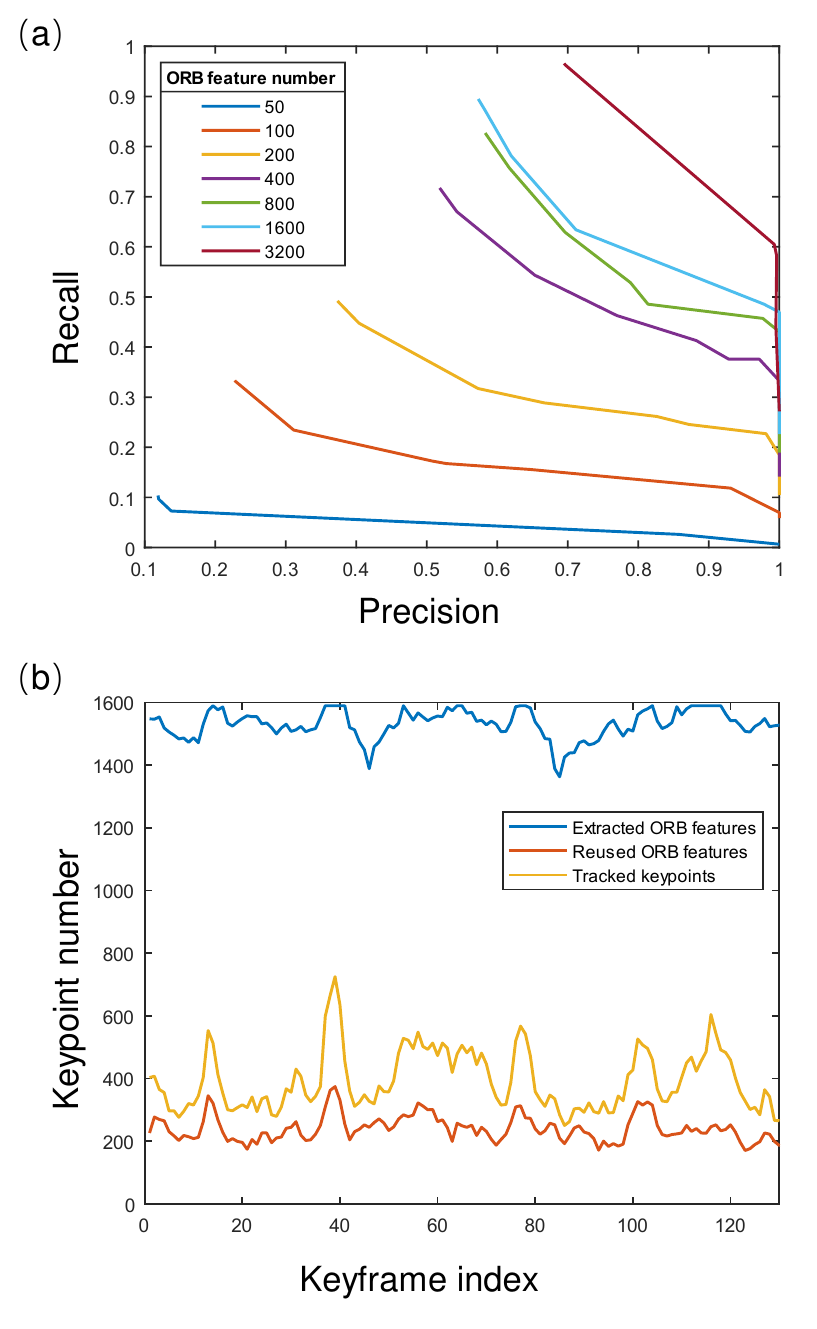}
\caption{(a) The precision-recall curve of loop closure detection with respect to different numbers of ORB features. The higher the curve, the better the performance of the algorithm. (b) Statistics of extracted ORB features (blue), reused ORB features (red) and all the keypoints tracked in the front-end (yellow) in one run.}
\label{fig:loop_test}
\end{figure}

The loop closure detection results with repect to different numbers of ORB features are shown in Fig. \ref{fig:loop_test}(a). 
It can be seen that with the increase of the number of ORB features from 100 to 3200, the recall rate at precision 100\% increases gradually, which proves that the performance of loop closure detection is positively correlated with the number of ORB features to a certain extent.
In order to hold a good trade-off between performance and speed, we set the ORB feature number to 1600 when running PA-SLAM.

Additionally, Fig. \ref{fig:loop_test}(b) shows the total number of extracted ORB features, the number of reused ORB keypoints fed into the depth filter and the number of all the tracked keypoints (including supplementary keypoints) when running PA-SLAM on this dataset. 
It can be seen that the ORB keypoints actually involved in the tracking front-end only account for about 15\% of all ORB features, which ensures the running efficiency of PA-SLAM.

\subsubsection{Loop closing in different travel directions}

\begin{figure*}[htb]
\centering\includegraphics{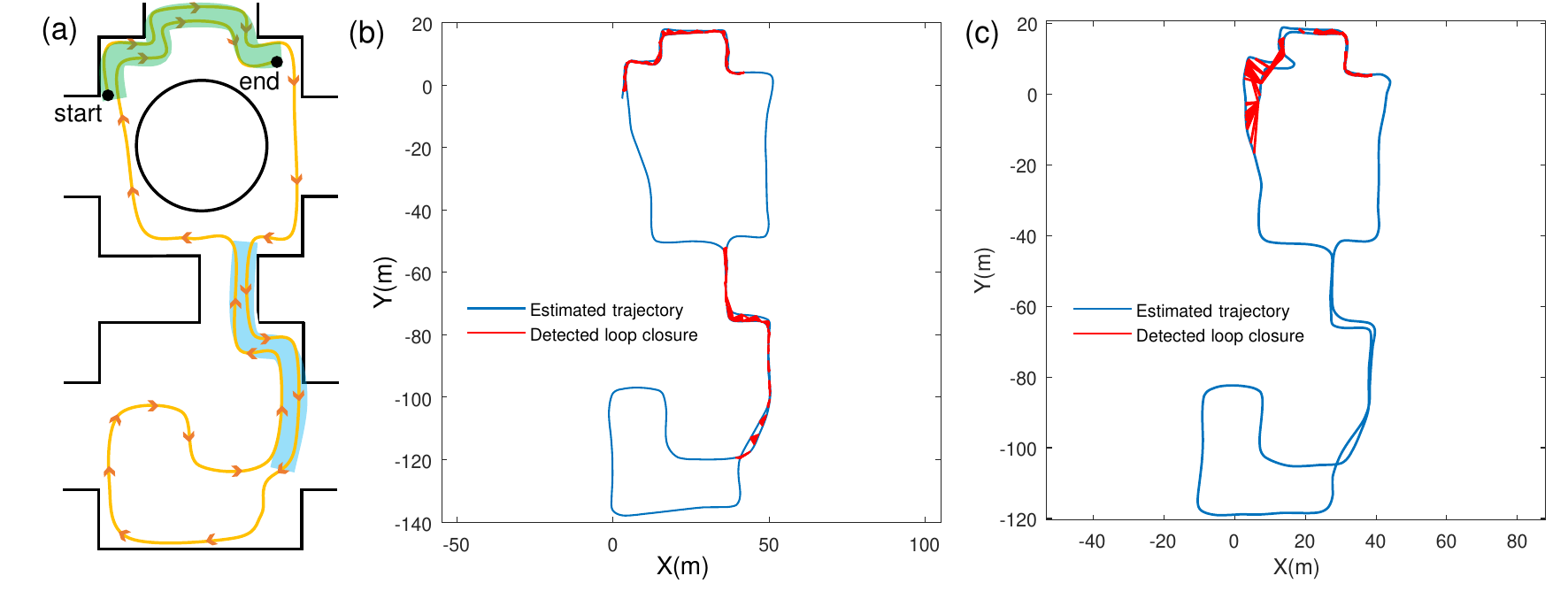}
\caption{(a) Schematic diagram of the path when collecting datasets for verifying the direction insensitivity of loop closure detection based on PAL. In this path, there exists the part traveling in the same direction (the green area) and the part traveling in the opposite direction (the blue area). (b) The estimate trajectory and results of loop closure detection produced by PA-SLAM. (c) The results of loop closure detection utilizing reconstructed perspective images.}
\label{fig:verse_loop_test}
\end{figure*}

In order to verify the insensitivity to the travel direction, we collect another video whose path is shown in Fig. \ref{fig:verse_loop_test}(a). 
There exists the part traveling in the same direction (the green area) and the part traveling in the opposite direction (the blue area) in this dataset. 
The estimated trajectory of PA-SLAM and successfully detected loop closure results (plot in red line segments) are shown in Fig. \ref{fig:verse_loop_test}(b). 
It can be seen that whether the travel direction is same or opposite, loop closure can always be correctly detected based on the PAL images, proving the direction insensitivity of loop closure in PA-SLAM.
As a contrast, the results of loop detection utilizing reconstructed perspective images are shown in Fig. \ref{fig:verse_loop_test}(c), indicating that only loop closure in the same travel direction can be detected successfully.

\subsection{Accuracy test}

\subsubsection{Accuracy test based on ArUco}

\begin{figure*}[htb]
\begin{center}
\includegraphics{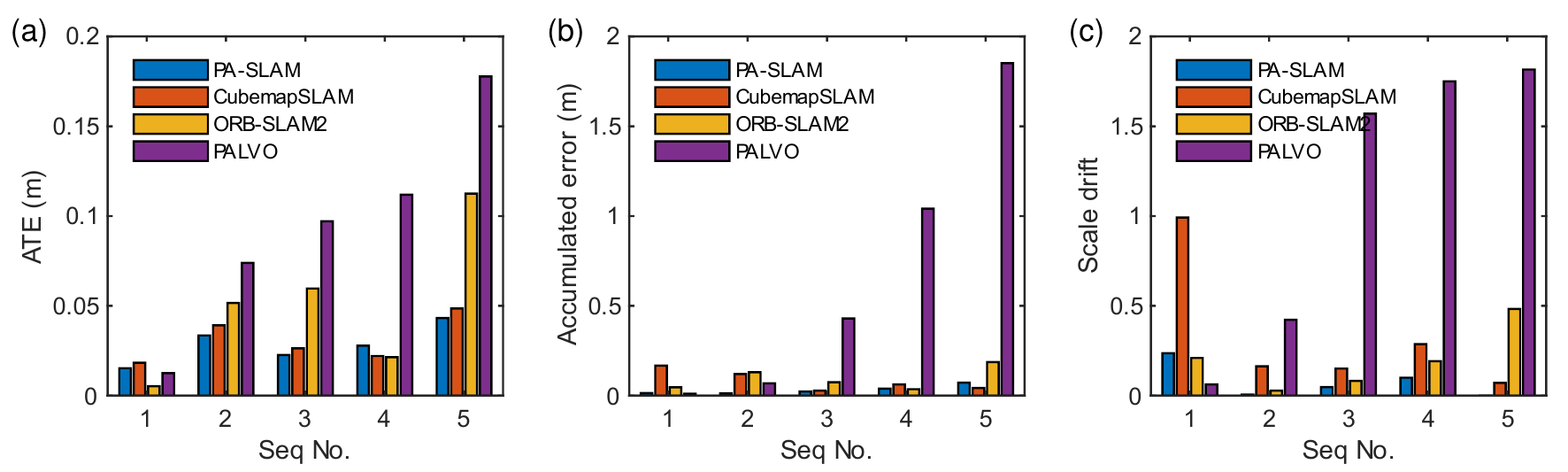}
\end{center}
\caption{Accuracy test results. We run PA-SLAM, CubemapSLAM, ORB-SLAM2 and PALVO on the datasets with ground truth, and calculate the (a) absolute trajectory error, (b) accumulated error and (c) scale drift separately. }
\label{fig:accuracy_test}
\end{figure*}

In this part, we evaluate the accuracy of PA-SLAM and compare it with the previous PALVO as well as CubemapSLAM \cite{cubemapslam},
which is a visual SLAM system based on panoramic images.
Simultaneously, comparative experiments with ORB-SLAM2 are also conducted, which is a state-of-the-art implementation of visual SLAM.
We use ArUco to obtain the ground truth of 6 degree of freedom (DOF) camera pose.
ArUco is an open-source library for camera pose estimation using squared markers \cite{garrido2016generation, romero2018speeded}.
The pixel correspondence necessary for pose estimation can be obtained through a single mark.
Thus, the camera pose can be calculated separately for each frame, and there is no error accumulation and scale drift over time.

Image sequences that are used in this test are captured in an office, with paths ranging from 3 meters to 50 meters in length.
It is impossible to capture the ArUco marker in all images in case of large scale camera movement. 
Thus, only part of the frames are assigned with ground truth. 
When collecting the datasets, we take the ArUco marker as the start point and the end point of the trajectory,
ensuring that frames in the beginning segment and the end segment have ground truth.

The absolute trajectory error (ATE) is utilized as the criterion for accuracy evaluation.
Additionally, the accumulated error and scale drift are also evaluated separately. 
Specifically, we align the tracked trajectory with the beginning segment (B) and the end segment (E) independently, providing two $Sim(3)$ transformations:

\begin{equation}
S_{b}^{\mathrm{gt}} =\underset{S \in \operatorname{Sim}(3)}{\operatorname{argmin}} \sum_{i \in B}\left(S p_{i}-p'_{i}\right)^{2},
\end{equation}

\begin{equation}
S_{e}^{\mathrm{gt}} =\underset{S \in \operatorname{Sim}(3)}{\operatorname{argmin}} \sum_{i \in E}\left(S p_{i}-p'_{i}\right)^{2}.
\end{equation}

The accumulated error ($e_{accu}$) and scale drift ($e_s$) can be defined as

\begin{equation}
e_{accu}=\sqrt{\frac{1}{n} \sum_{i=1}^{n}\left\|S_{b}^{\mathrm{gt}} p_{i}-S_{e}^{\mathrm{gt}} p_{i}\right\|_{2}^{2}},
\end{equation}

\begin{equation}
    e_s = \left| log( scale( S_{b}^{\mathrm{gt}} ( S_{e}^{\mathrm{gt}} ) ^ {-1} ) ) \right|.
\end{equation}

Fig. \ref{fig:accuracy_test} presents the experiment results, from which one can see that our algorithm achieves the least ATE on the sequence (2), (3) and (5),
while on the sequence (1) and (4) ORB-SLAM2 performs best w.r.t. ATE.
As for the accumulated error, our PA-SLAM delivers superior performance on the sequence (1)-(4), but slightly inferior to CubemapSLAM on the sequence (5).
For scale drift, PA-SLAM achieves the best performance among the four algorithms on the sequences (2)-(5).
There is an exception of sequence (1), on which PALVO performs better.
This is because the movement scale of this sequence is quite small (the path length of sequence (1) is about 3 meters).
Under this circumstance, PALVO maintains good local consistency of the trajectory, with error accumulation and scale drift not being significant.

The experiment results indicate that the proposed PA-SLAM has achieved equivalent or even better accuracy in comparison with ORB-SLAM2 and CubemapSLAM, and has been greatly improved compared to the previous PALVO.
It becomes clear that loop closure and global optimization significantly decrease error accumulation and scale drift in large-scale and long-term running.

\subsubsection{Accuracy test of loop closure error}

In addition, we also run our algorithm on the dataset used in the accuracy test of PALVO to collect and compare the overall numerical performance. 
As described in the paper of PALVO, this dataset is collected in an indoor corridor and contains a total of 5 videos (R1 - R5), with paths ranging from 20 meters to 50 meters in length.
The start and end point are exactly in the same position.
Loop closure error in percentage is utilized as a criterion for accuracy evaluation,
which is defined as the ratio of the residual between the start- and end points of the trajectory estimated by algorithms, to the whole length of estimated trajectory:
\begin{equation}
    e_{loop} = 
    \frac{\|P_{start}-P_{end}\|}{L_{traj}} \times 100\%.
\label{eq:loop_closure_error}
\end{equation}

Table~\ref{tab:accuracy_test} presents the quantitative results.
As can be seen, the proposed PA-SLAM achieves the least loop closure error in R1, R3 and R4. 
In R2 PA-SLAM is inferior to CubemapSLAM,
and in R5 it is slightly inferior to ORB-SLAM2  but still better than the other three algorithms.
These experiment results further support our conclusion that PA-SLAM reaches the state-of-the-art performance and has a great improvement compared to the previous PALVO.

\begin{table*}[ht]
\centering
\caption{\bf Accuracy test results.}
\label{tab:accuracy_test}
\begin{tabular}{llclccccc}
\hline
          &  & Frame rate     &  & \multicolumn{5}{c}{Loop closure error (\%)}                                             \\ \hline
Method    &  & FPS            &  & R1              & R2              & R3              & R4              & R5              \\ \cline{1-1} \cline{3-3} \cline{5-9} 
PA-SLAM   &  & 99.1           &  & \textbf{0.6749} & 1.1702          & \textbf{0.4417} & \textbf{0.9060} & 0.7553          \\
CubemapSLAM& & 25.3           &  & 0.7781          & \textbf{0.4219} & 0.7647          & 0.9553          & 0.8439          \\
ORB-SLAM2 &  & 37.4           &  & 0.8364          & 1.2000          & 0.5425          & 1.2681          & \textbf{0.6779} \\
PALVO     &  & 251.6          &  & 1.9326          & 1.5893          & 2.9858          & 2.6105          & 0.9527          \\ 
SVO       &  & \textbf{423.9} &  & 1.4276          & 2.0067          & 2.8455          & 3.5414          & 1.7109          \\ \hline
\end{tabular}
\end{table*}

Moreover, we also evaluate the frame rate of our algorithm.
With loop closing and global optimization, the proposed PA-SLAM is capable of processing frames at 99.1 frames per second (FPS), which is much faster than ORB-SLAM2 and CubemapSLAM.

\subsection{Field test}

\begin{table*}[ht]
\centering
\caption{\bf Field test results.}
\label{tab:field_test}
\setlength{\tabcolsep}{4mm}{
\begin{tabular}{llcccc}
\hline
            &  & \multicolumn{4}{c}{Loop closure error (\%)}                                \\ \hline
Method      &  & S1 (190 m)      & S2 (450 m)       & S3 (200 m)        & S4 (250 m)        \\ \cline{1-1} \cline{3-6} 
PA-SLAM     &  & \textbf{0.5361} & \textbf{0.0717}  & \textbf{0.1136}   & \textbf{0.2789}   \\
CubemapSLAM &  & 0.9989          & 2.4388           & 4.3081            & 4.0793            \\
ORB-SLAM2   &  & 1.2254          & -                & 6.3477            & -                 \\
PALVO       &  & 2.5719          & 4.2268           & 4.2288            & 3.6310            \\ 
SVO         &  & 3.6702          & -                & -                 & 6.5322            \\ \hline
\end{tabular}
}
\end{table*}

\begin{figure*}[!t]
\centering
\includegraphics{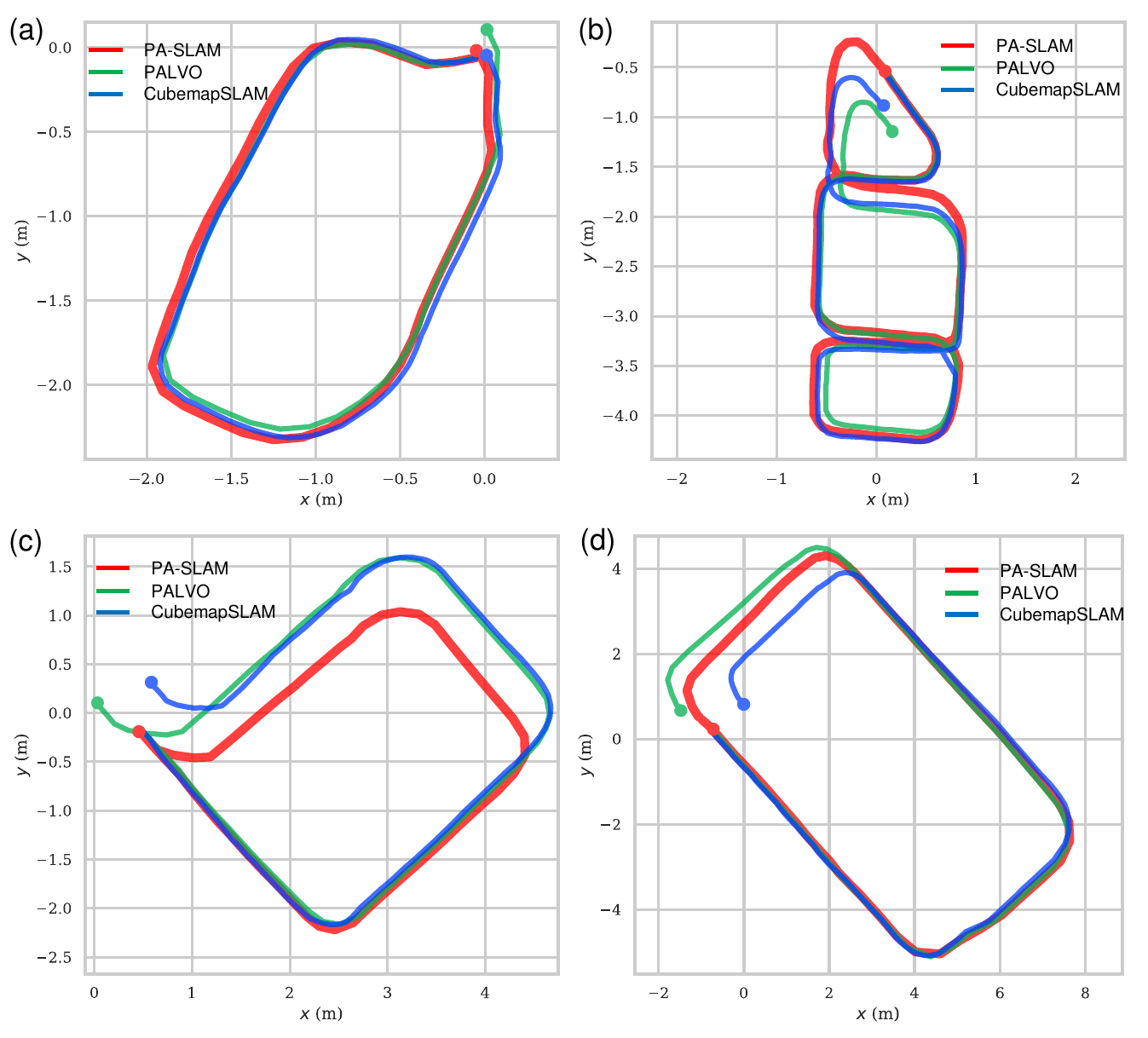}
\caption{ Trajectories produced by PA-SLAM, PALVO and CubemapSLAM on datasets (a) S1 - (d) S4. The start points of different estimated trajectories are aligned to the same point, and the end points of each trajectory are indicated by dots in different colors. In (a) the orientation of the remote control vehicle at the start point is approximately perpendicular to the end point, which is opposite to the end point in (c) and the same as the end point in (b)(d). It’s worth noting that the trajectories are plotted up to a scale factor, because the absolute scale cannot be derived from a single camera.}
\label{fig:field_test}
\end{figure*}

\begin{figure*}[!t]
\begin{center}
\includegraphics{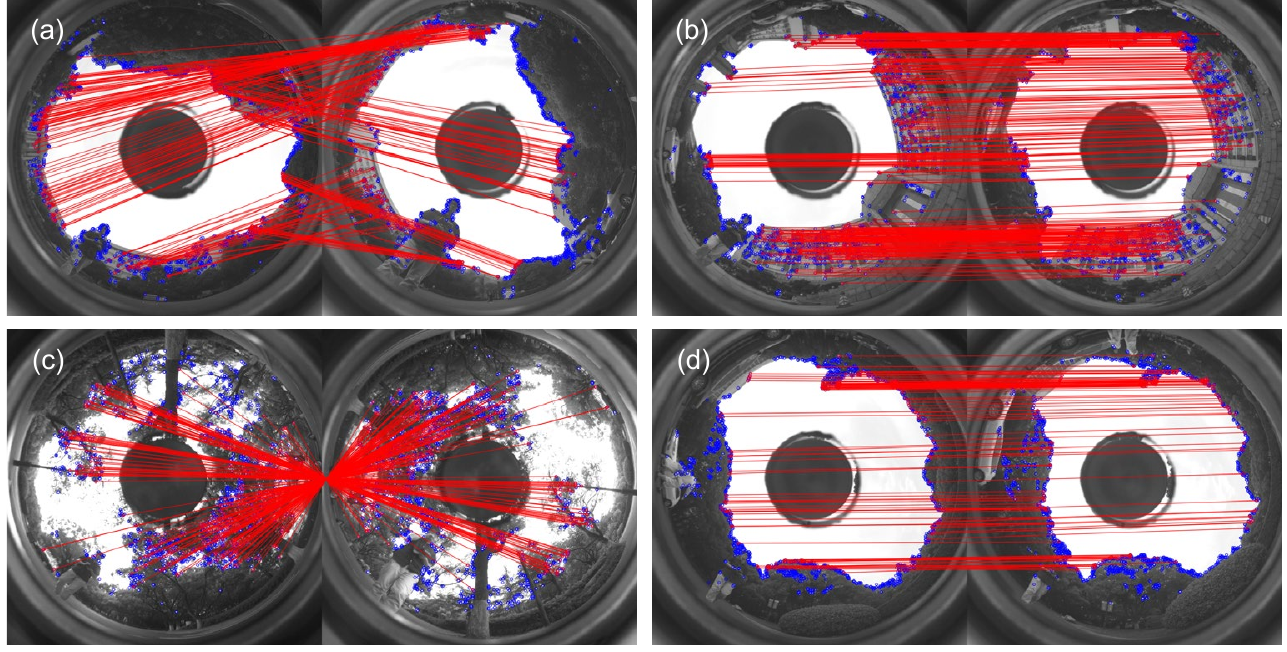}
\end{center}
\caption{Feature matching when the vehicle revisits a certain place (a loop closure occurs) with its orientation (a) perpendicular to-, (c) opposite to- and (b)(d) the same as the first visit. }
\label{fig:feat_matching}
\end{figure*}

In order to further verify our algorithm and validate its effectiveness and reliability in real applications, field tests are conducted in the outdoor area.
We collect a number of videos in the campus, ranging from 190 to 450 meters in length.
In these videos, there are a large number of pedestrians, vehicles and other dynamic components, which is challenging for conventional visual SLAM systems.
Similarly, the start- and end point are kept in the same place and the loop closure error in percentage is calculated.
Table \ref{tab:field_test} displays the experiment results, and the estimated trajectories are shown in Fig. \ref{fig:field_test}. 

As can be seen in Table \ref{tab:field_test}, PA-SLAM achieves least loop closure errors on all of the sequences.
Additionally, the perspective image-based ORB-SLAM2 and SVO get failed on two of the sequences.
Fig. \ref{fig:field_test} depicts the trajectories produced by PA-SLAM, PALVO and CubemapSLAM.
from which one can see that the orientation of the remote control vehicle at the start point is approximately perpendicular to the end point in S1, and opposite to the end point in S3.
In spite of this issue, PA-SLAM can still close the loop, further proving the direction insensitivity of loop closure in PA-SLAM.
Fig. \ref{fig:feat_matching} represents the ORB feature matching when the vehicle revisits a certain place (a loop closure occurs) with its orientation perpendicular to-, opposite to- and the same as the first visit, demonstrating the robustness of PA-SLAM in real-world unconstrained scenarios.

\section{Conclusion}

In this paper, we propose PA-SLAM, which extends the sparse direct method based PALVO to PA-SLAM with loop closure detection and global optimization.
The hybrid point selection is presented to enable reliable BoW-based loop closure detection while ensuring computational efficiency.
When a loop closure is successfully detected, pose graph optimization is performed and followed by global BA.
Experiments demonstrate that PA-SLAM significantly reduces the error accumulation and scale drift in PALVO, reaching state-of-the-art accuracy and maintaining the original robustness and high efficiency. 
Meanwhile, PA-SLAM can deal with loop closure in different travel directions, which greatly improves the performance in practical application scenarios.

\begin{backmatter}
\bmsection{Funding}
This research was granted from ZJU-Sunny Photonics Innovation Center (No. 2020-03). This research was also funded in part through the AccessibleMaps project by the Federal Ministry of Labor and Social Affairs (BMAS) under the Grant No. 01KM151112.

\bmsection{Acknowledgments}
This research was supported in part by Hangzhou SurImage Technology Company Ltd.

\bmsection{Disclosures}
The authors declare no conflicts of interest.

\bmsection{Data availability} Data underlying the results presented in this paper are not publicly available at this time but may be obtained from the authors upon reasonable request.

\end{backmatter}

\bibliography{ref}

% Full bibliography added automatically for Optics Letters submissions; the following line will simply be ignored if submitting to other journals.
% Note that this extra page will not count against page length
\bibliographyfullrefs{ref}

%Manual citation list
%\begin{thebibliography}{1}
%\bibitem{Zhang:14}
%Y.~Zhang, S.~Qiao, L.~Sun, Q.~W. Shi, W.~Huang, %L.~Li, and Z.~Yang,
 % \enquote{Photoinduced active terahertz metamaterials with nanostructured
  %vanadium dioxide film deposited by sol-gel method,} Opt. Express \textbf{22},
  %11070--11078 (2014).
%\end{thebibliography}

% Please include bios and photos of all authors for aop articles
\ifthenelse{\equal{\journalref}{aop}}{%
\section*{Author Biographies}
\begingroup
\setlength\intextsep{0pt}
\begin{minipage}[t][6.3cm][t]{1.0\textwidth} % Adjust height [6.3cm] as required for separation of bio photos.
  \begin{wrapfigure}{L}{0.25\textwidth}
    \includegraphics[width=0.25\textwidth]{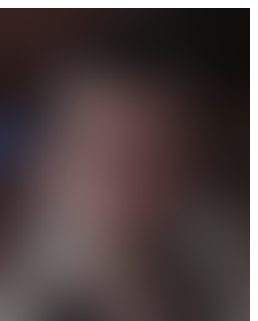}
  \end{wrapfigure}
  \noindent
  {\bfseries John Smith} received his BSc (Mathematics) in 2000 from The University of Maryland. His research interests include lasers and optics.
\end{minipage}
\begin{minipage}{1.0\textwidth}
  \begin{wrapfigure}{L}{0.25\textwidth}
    \includegraphics[width=0.25\textwidth]{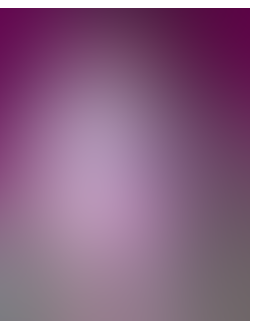}
  \end{wrapfigure}
  \noindent
  {\bfseries Alice Smith} also received her BSc (Mathematics) in 2000 from The University of Maryland. Her research interests also include lasers and optics.
\end{minipage}
\endgroup
}{}

\end{document}